\documentclass{pnastwo}

\usepackage[pdftex]{graphicx}
\graphicspath{ {images/} }
\usepackage{float}
\usepackage{listings}
\usepackage[utf8]{inputenc}
\usepackage{lipsum}  
\usepackage{biblatex}
\usepackage{caption}
\usepackage{subcaption}
\usepackage{enumitem}
\usepackage{csquotes}
\usepackage{hyperref}
\hypersetup{
    colorlinks=true,
    linkcolor=blue,
    filecolor=magenta,      
    urlcolor=blue,
}

\usepackage{amssymb,amsfonts,amsmath}

\contributor{Published as a whitepaper}
\copyrightyear{2016}
\issuedate{August 2016}
\volume{Vitoria}
\issuenumber{Spain}

\begin{document}




\title{Extending the OpenAI Gym for robotics: a toolkit for reinforcement learning using ROS and Gazebo}






\author{
Iker Zamora\affil{1}{Erle Robotics},
Nestor Gonzalez Lopez\affil{1}{Erle Robotics},
Víctor Mayoral Vilches\affil{1}{Erle Robotics},
\and
Alejandro Hernández Cordero\affil{1}{Erle Robotics}}


\maketitle

\begin{article}

\begin{abstract}
This paper presents an extension of the OpenAI Gym for  robotics using the Robot Operating System (ROS) and the Gazebo simulator. The content discusses the software architecture proposed and the results obtained by using two Reinforcement Learning techniques: Q-Learning and Sarsa. Ultimately, the output of this work presents a benchmarking system for robotics that allows different techniques and algorithms to be compared using the same virtual conditions.
 \end{abstract}






\section{Introduction}

Reinforcement Learning (RL) is an area of machine learning where a software agent learns by interacting with an environment, observing the results of these interactions with the aim of achieving the maximum possible cumulative reward. This imitates the trial-and-error method used by humans to learn, which consists of taking actions and receiving positive or negative feedback.\\
\newline
In the context of robotics, reinforcement learning offers a framework for the design of sophisticated and hard-to-engineer behaviors \cite{kober2013reinforcement}.  The challenge is to build a simple environment where this machine learning techniques can be validated, and later applied in a real scenario.\\
\newline
OpenAI Gym \cite{brockman2016openai} is a is a toolkit for reinforcement learning research that has recently gained popularity in the machine learning community. The work presented here follows the same baseline structure displayed by researchers in the OpenAI Gym  (\href{https://gym.openai.com}{gym.openai.com}), and builds a gazebo environment on top of that. OpenAI Gym focuses on the episodic setting of RL, aiming to maximize the expectation of total reward each episode and to get an acceptable level of performance as fast as possible. This toolkit aims to integrate the Gym API with robotic hardware, validating reinforcement learning algorithms in real environments. Real-world operation is achieved combining Gazebo simulator \cite{koenig2006gazebo}, a 3D modeling and rendering tool, with ROS \cite{quigley2009ros} (Robot Operating System), a set of libraries and tools that help software developers create robot applications.\\
\newline
As benchmarking in robotics remains an unsolved issue, this work aims to provide a toolkit for robot researchers to compare their techniques in a well-defined (API-wise) controlled environment that should speed up development of  robotic solutions.




\section{Background}

Reinforcement Learning has taken an increasingly important role for its application in robotics \cite{rlintro}. This technique offers robots the ability to learn previously missing abilities \cite{kormushev2013reinforcement} like learning hard to code behaviours or optimizing problems without an accepted \emph{closed} solution.\\
\newline
The main problem with RL in robotics is the high cost per-trial, which is not only the economical cost but also the long time needed to perform learning operations. Another known issue is that learning with a real robot in a real environment can be dangerous, specially with flying robots like \emph{quad-copters}. In order to overcome this difficulties, advanced robotics simulators like Gazebo have been developed which help saving costs, reducing time and speeding up the simulation.\\
\newline
The idea of combining learning in simulation and in a real environment was popularized by the Dyna-architecture (Sutton, 1990), prioritized sweeping (Moore and Atkeson,1993), and incremental multi-step Q-Learning (Pengand Williams, 1996) in reinforcement learning. In robot reinforcement learning, the learning step on the simulated system is often called “mental rehearsal” \cite{kober2013reinforcement}. Mental reharsal in robotics can be improved by obtaining information from the real world in order to create an accurate simulated environment. Once the training is done, just the resulting policy is transferred to the real robot.\\


\section{Architecture}

\begin{figure}[H!]
    \centering
    \includegraphics[width=0.48\textwidth]{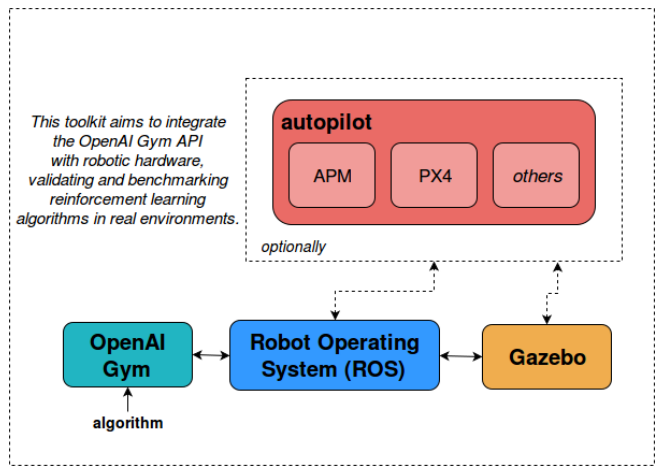}
    \caption{Simplified software architecture used in OpenAI Gym for robotics. 
    \label{fig:diagram}}
\end{figure}

The architecture consits of three main software blocks: OpenAI Gym, ROS and Gazebo (Figure \ref{fig:diagram}). Environments developed in OpenAI Gym interact with the Robot Operating System, which is the connection between the Gym itself and Gazebo simulator. Gazebo provides a robust physics engine, high-quality graphics, and convenient programmatic and graphical interfaces. \\
\newline
The physics engine needs a robot definition\footnote{Unified Robot Description Format (URDF)} in order to simulate it, which is provided by ROS or a gazebo plugin that interacts with an autopilot in some cases (depends on the robot software architecture). The Turtlebot is encapsulated in ROS packages while robots using an autopilot like Erle-Copter and Erle-Rover are defined using the corresponding autopilot. Our policy is that every robot needs to have an interface with ROS, which will mantain an organized architecture.\\
\newline


Figure \ref{fig:diagram} presents the a simplified diagram of the software architecture adopted in our work. Our software structure provides similar APIs to the ones presented initially by OpenAI \emph{Gym}. We added a new collection of environments called \emph{gazebo} where we store our own gazebo environments with their corresponding assets. The needed installation files are stored inside the \emph{gazebo} collection folder, which gives the end-user an easier to modify infrastructure.\\
\newline
Installation and setup consits of a ROS catkin workspace containing the ROS packages required for the robots (e.g.: Turtlebot, Erle-Rover and Erle-Copter) and optionally the appropriate autopilot that powers the logic of the robot. In this particular case we used the APM autopilot thereby the source code is also required for simulating Erle-Rover and Erle-Copter. 
Robots using APM stack need to use a specific plugin in order to communicate with a ROS/Gazebo simulation.

\section{Environments and Robots}

\begin{figure}[H!]
    \centering
    \includegraphics[width=0.48\textwidth]{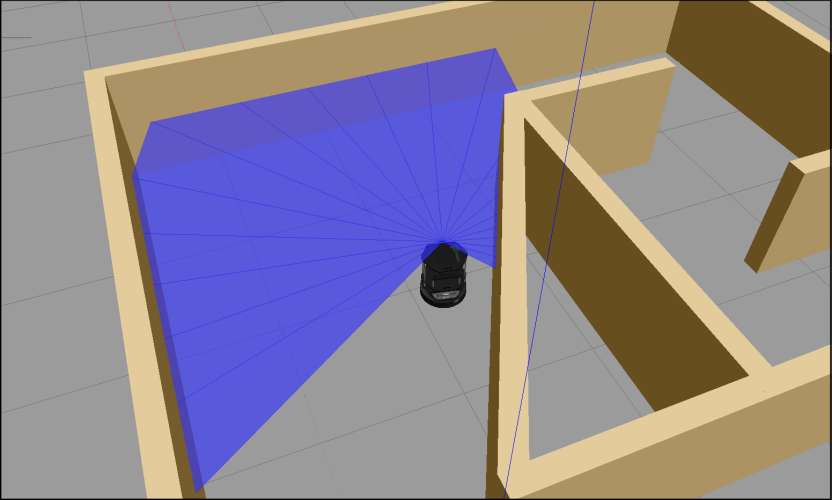}
    \caption{Turtlebot in a virtual environment learning how to navigate autonomously using a LIDAR \label{fig:turtlesilly}}
\end{figure}

We have created a collection of six environments for three robots: Turtlebot, Erle-Rover and Erle-Copter. Following the desing decisions of OpenAI Gym, we only provide an abstraction for the environment, not the agent. This means each environment is an independent set of items formed mainly by a \emph{robot} and a \emph{world}. \\
\newline 
Figure \ref{fig:turtlesilly} displays an environment created with the Turtlebot \emph{robot} which has been provided with a LIDAR sensor using and a \emph{world} called \emph{Circuit}. If we wanted to test our reinforcement learning algorithm with the Turtlebot but this time using positioning information, we would need to create a completely new environment.\\
\newline

The following are the initial environments and robots provided by our team at Erle Robotics. Potentially, the amount of supported robots/environments will will grow over time.\\
\newline

\begin{figure}[h!]
    \centering
    \begin{subfigure}[t]{0.2\textwidth}
        \captionsetup{width=0.95\linewidth}
        \centering
        \includegraphics[width=0.95\linewidth]{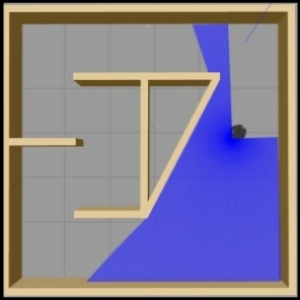}
        \caption{A simple circuit with a diagonal wall, which increases the complexity of the learning.\label{fig:circuit1}}
    \end{subfigure}%
    \begin{subfigure}[t]{0.2\textwidth}
        \captionsetup{width=0.95\linewidth}
        \centering
        \includegraphics[width=0.95\linewidth]{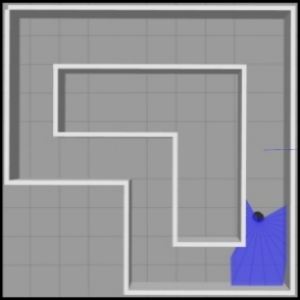}
        \caption{A simple circuit with straight tracks and 90 degree turns. Note that the third curve is a left turn, while the others are right turns. \label{fig:circuit2}}
    \end{subfigure}%

    \begin{subfigure}[t]{0.2\textwidth}
        \captionsetup{width=0.95\linewidth}
        \centering
        \includegraphics[width=0.95\linewidth]{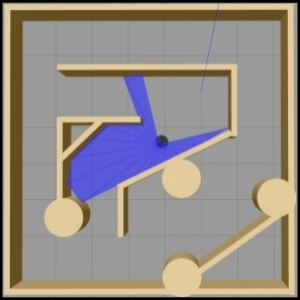}
        \caption{A complex maze with different wall shapes and some narrow tracks.\label{fig:maze}}
    \end{subfigure}%
    \begin{subfigure}[t]{0.2\textwidth}
        \captionsetup{width=0.95\linewidth}
        \centering
        \includegraphics[width=0.95\linewidth]{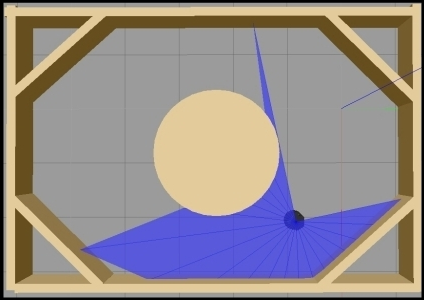}
        \caption{A simple oval shaped circuit.\label{fig:round}}
    \end{subfigure}
    \caption{Environments available using Turtlebot robot and a LIDAR sensor. \label{fig:turtlebotEnvs}}
\end{figure}

\textbf{Turtlebot.} TurtleBot combines popular off-the-shelf robot components like the iRobot Create, Yujin Robot's Kobuki, Microsoft's Kinect and Asus' Xtion Pro into an integrated development platform for ROS applications. For more information, please see \href{http://turtlebot.com}{turtlebot.com} .\\
\newline
The following are the four environment currently available for the Turtlebot:
\begin{itemize}[leftmargin=.2in]
    \item \emph{GazeboCircuitTurtlebotLIDAR-v0} (Figure \ref{fig:turtlebotEnvs}.a) : A simple circuit with a diagonal wall, which increases the complexity of the learning.
    \item \emph{GazeboCircuit2TurtlebotLIDAR-v0} (Figure \ref{fig:turtlebotEnvs}.b) : A simple circuit with straight tracks and 90 degree turns. Note that the third curve is a left turn, while the others are right turns.
    \item \emph{GazeboMazeTurtlebotLIDAR-v0} (Figure \ref{fig:turtlebotEnvs}.c) : A complex maze with different wall shapes and some narrow tracks.
    \item \emph{GazeboRoundTurtlebotLIDAR-v0} (Figure \ref{fig:turtlebotEnvs}.d) : A simple oval shaped circuit.
\end{itemize}
\textbf{Erle-Rover.} A Linux-based smart car powered by the APM autopilot and with support for the Robot Operating System \href{http://erlerobotics.com/blog/erle-rover/}{erlerobotics.com/blog/erle-rover} .\\
\newline
\begin{figure}[ht!]
    \centering
    \includegraphics[width=0.48\textwidth]{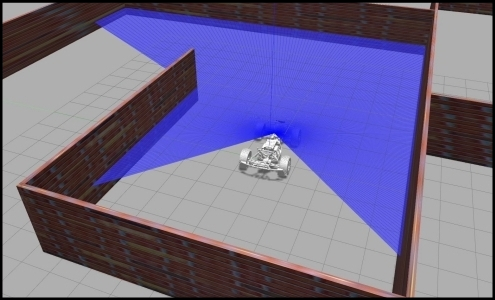}
    \caption{Erle-Rover learning to avoid obstacles in a big maze with wide tracks and 90 degree left and right turns, an environment called \emph{GazeboMazeErleRoverLIDAR-v0}.\label{fig:apm1}}
\end{figure}
\textbf{Erle-Copter.} A Linux-based drone powered by the open source APM autopilot and with support for the Robot Operating System \href{http://erlerobotics.com/blog/erle-copter/}{erlerobotics.com/blog/erle-copter} .
\begin{figure}[ht!]
    \centering
    \includegraphics[width=0.48\textwidth]{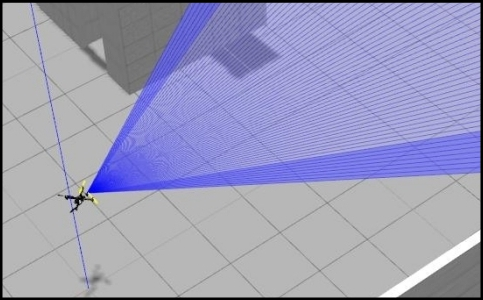}
    \caption{Erle-Copter learning to avoid obstacles in ErleRobotics office without ceiling, an environment called \emph{GazeboOfficeErleCopterLIDAR-v0}.\label{fig:apm2}}
\end{figure}


\section{Results}
We have experimented with two Reinforcement Learning algorithms, Q-Learning and Sarsa. The turtlebot has been used to benchmark the algorithms since we get faster simulation speeds than robots using and autopilot. We get around 60RTF (Real Time Factor), which means 60 times normal simulation speed and 30RTF when we launch the visual interfaze. This benchmarks have been made using a i7 6700 CPU and non-GPU laser mode. 
\begin{figure}[ht!]
    \centering
    \includegraphics[width=0.48\textwidth]{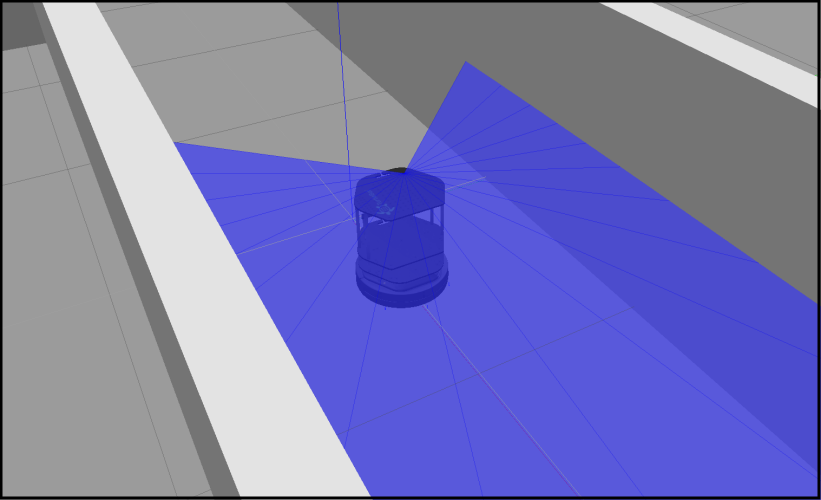}
    \caption{Turtlebot with LIDAR in \it{GazeboCircuit2TurtlebotLIDAR-v0} environment.\label{fig:c2result}}
\end{figure}
We will use \emph{GazeboCircuit2TurtlebotLIDAR-v0} (Figure \ref{fig:turtlebotEnvs}.b \& \ref{fig:c2result}) to perform the benchmarking. This environment consists of a simple straight lined circuit with five right turns and one left turn. We will use just a LIDAR sensor to train the Turtlebot, no positioning or other kind of data will be used. Both algorithms will use the same hyperparameters and exact environment. The actions and rewards forming the environment have been adapted to get an optimal training performance.\\
\newline
\textbf{Actions}
\begin{itemize}[leftmargin=.2in]
\item Forward: v = 0.3 m/s
\item Left/Right: v = 0.05 m/s , w = +-0.3 rad/s
\end{itemize}
Taking only three actions will lead to a faster learning, as the 'Q' function will fill its table faster. Obstacle avoidance could be performed with just two turning actions, but the learnt movements would be less practical.\\
\newline
Left and right turns have a small linear velocity just to accelerate the learning process and avoid undesired behaviours. Setting the linear velocity to zero, the robot could learn to turn around itself constantly as it would not crash and still earn positive rewards. As Atkeson and Schaal point out:
\begin{displayquote}
Reinforcement learning approaches exploit such model inaccuracies if they are beneficial for the reward received in simulation.
\end{displayquote}
This could be avoided changing the reward system, but we found the optimal values for this environment are the following.\\
\newline
\textbf{Rewards}
\begin{itemize}[leftmargin=.2in]
\item Forward: 5
\item Left/Right: 1
\item Crash: -200
\end{itemize}
Forward actions take five times more reward than turns, this will make the robot take more forward actions as they give more reward. We want to take as many forward actions as possible so that the robot goes forward in straight tracks, which will lead to a faster and more realistic behaviour.\\
\newline
Left and right actions are rewarded with 1, as they are needed to avoid crashes too. Setting them higher would result in a zigzagging behaviour.\\
\newline
Crashes earn very negative rewards for obvious reasons, we want to avoid obstacles.\\
\newline
\textbf{Q-Learning}\label{qlearning}\\
\newline
Q-Learning\cite{watkins1992q} is an Off-Policy algorithm for Temporal Difference learning. Q-Learning learns the optimal policy even when actions are selected according to an exploratory or even random policy \cite{sarsavsqlearn2}.
$$Q(s_t,a_t) \leftarrow Q(s_t,a_t) + \alpha [r_{t+1} + \gamma max{\alpha}, Q(s_{t+1}, a_{t})-Q(s_t,a_t)]$$

Let's see how the Turtlebot learns using Q-Learning in \emph{GazeboCircuit2TurtlebotLIDAR-v0} environment. Simplified code is presented below:\\

\begin{lstlisting}[basicstyle=\footnotesize, language=Python]
env = gym.make('GazeboCircuit2TurtlebotLIDAR-v0')
qlearn = qlearn.Qlearn(alp=0.2,gam=0.9,eps=0.9)
for x in range(3000):
   observation = env.reset()
   state = ''.join(map(str, observation))
   for i in range(1500):
      action = sarsa.chooseAction(state)
      observation, reward, done = env.step(action)
      nextState = ''.join(map(str, observation))
      qlearn.learn(state, action, reward, 
                  nextState)
      if not(done):
         state = nextState
      else:
         break 
\end{lstlisting}
After selecting environment we want to test, we have to initialize Q-learn with three parameters. Small changes in these hyperparameters can result in substantial changes in the learning of our robot. Those parameters are the following:\\
\newline
\begin{itemize}[leftmargin=.2in]
\item \underline{$\alpha$, Learning rate}:. Setting it to 0 means the robot will not learn and a high value such as 0.9 means that learning can occur quickly.
\item \underline{$\gamma$, Discount factor}:. A factor of 0 will make the agent consider only current rewards, while a factor approaching 1 will make it strive for a long-term high reward.
\item \underline{$\epsilon$, Exploration constant}:. Used to randomize decisions, setting a high value such as 0.9 will make 90\% of the actions to be stochastic. An intersting technique is to set an epsilon decay, where the agent starts taking more randon actions (exploration phase) and ends in an exploitation phase where all or most of the actions performed are selected from the learning table instead of being random.
\end{itemize}
The selected initial hyperparameters are, $\alpha=0.2$, $\gamma=0.9$ and $\epsilon=0.9$. In this example we use the epsilon decay technique, being the decay $\epsilon*0.9986$ every episode until it reaches our minimum epsilon value, 0.05 in this case.\\
\newline
We want to run the simulation for 3000 episodes. Each episode the simulation will be resetted and the robot will start again from its initial position. Every episode we try to make a maximum of 1500 iterations, which means the robot has not crashed. Every iteration we choose an action, take a step (execute an action for a short time or distance) and receive feedback. That feedback is called \emph{observation} and it is returns the next state to be taken and the received reward.\\
\newline
To sum up, every episode the robot tries to take as many steps as possible, learning every step from the obtained rewards.\\
\newline
The following graph shows the results obtained through 3000 episodes. 

\begin{figure}[h!]
    \centering
    \includegraphics[width=0.48\textwidth]{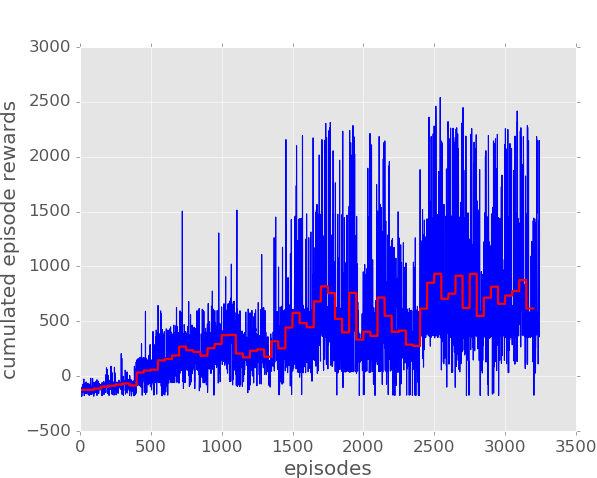}
    \caption{Cumulated reward graph obtained from the monitoring of \emph{GazeboCircuit2TurtlebotLIDAR-v0} (Figure \ref{fig:turtlebotEnvs}.b) environment using Q-Learning. The blue line  prints the whole set of readings while the red line shows an approximation to the averaged rewards. \label{fig:qlearnplot}}
\end{figure}

We get decent results after 1600 episodes. Cumulated rewards around 2000 or higher usually mean the robot did not crash or gave more than two laps. \\
\newline
\textbf{Sarsa}\\
\newline
Sarsa \cite{rummery1994line} is an On-Policy algorithm for Temporal Difference Learning. The major difference between it and Q-Learning, is that the maximum reward for the next state is not necessarily used for updating the Q-values. Instead, a new action, and therefore reward, is selected using the same policy that determined the original action  \cite{sarsavsqlearn2}.
$$Q(s_t,a_t) \leftarrow Q(s_t,a_t) + \alpha [r_{t+1} + \gamma Q(s_{t+1}, a_{t+1})-Q(s_t,a_t)]$$

Let's now compare how the Turtlebot learns using Sarsa in \emph{GazeboCircuit2TurtlebotLIDAR-v0} environment. Simplified code is presented below\\

\begin{lstlisting}[basicstyle=\footnotesize, language=Python]
env = gym.make('GazeboCircuit2TurtlebotLIDAR-v0')
sarsa = sarsa.Sarsa(alp=0.2,gam=0.9,eps=0.9)
for x in range(3000):
   observation = env.reset()
   state = ''.join(map(str, observation))
   for i in range(1500):
      action = sarsa.chooseAction(state)
      observation, reward, done = env.step(action)
      nextState = ''.join(map(str, observation))
      nextAction = sarsa.chooseAction(nextState)
      sarsa.learn(state, action, reward, 
                  nextState, nextAction)
      if not(done):
         state = nextState
      else:
         break 
\end{lstlisting}
After selecting environment we want to test, we have to initialize Sarsa with three parameters: alpha, gamma and epsilon. They work in the same way as Q-learn, so we used the same as in the previous Q-leaning test \ref{qlearning}. The selected initial hyperparameters are, $\alpha=0.2$, $\gamma=0.9$ and $\epsilon=0.9$ with $\epsilon*0.9986$ epsilon decay. \\
\newline
We want to run the simulation for 3000 episodes. Every episode we try to make a maximum of 1500 iterations, which means the robot has not crashed. Every iteration we choose an action, take a step (execute an action for a short time or distance) and receive feedback. That feedback is called \emph{observation} and it is used to build the next action to be taken.\\
\newline
Since we are using Sarsa (on-policy) and not Q-learn (off-policy), we need to choose another action before we learn. This is done by selecting a new action from the previously built next state.\\
\newline
As explained in the Q-Learning section, every episode the robot tries to take as many steps as possible, learning every step from the obtained rewards.\\
\newline
The following graph shows the results obtained through 3000 episodes.
\begin{figure}[h!]
    \centering
    \includegraphics[width=0.48\textwidth]{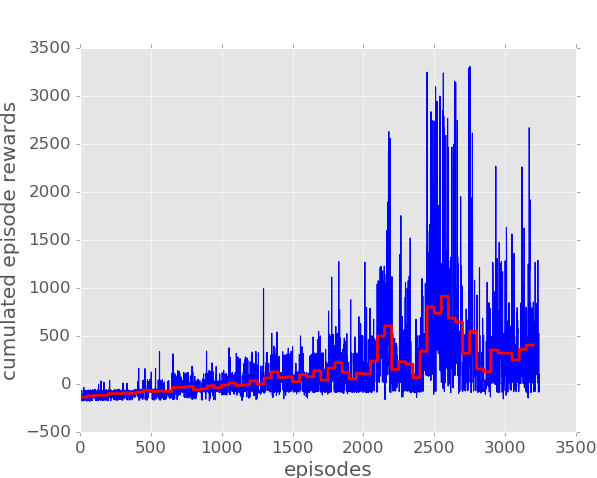}
    \caption{Cumulated reward graph obtained from the monitoring of \emph{GazeboCircuit2TurtlebotLIDAR-v0} (Figure \ref{fig:turtlebotEnvs}.b) environment using Sarsa. The blue line  prints the whole set of readings while the red line shows an approximation to the averaged rewards.\label{fig:sarsaplot}}
\end{figure}
We get decent results after 2500 episodes. Cumulated rewards around 2000 or higher usually mean the robot did not crash or gave more than two laps.\\
\newline
\textbf{Benchmarking}\\
\newline
The learning in Q-Learning occurs faster than in Sarsa, this happens because Q-Learning is able to learn a policy even if taken actions are chosen randomly. However, Q-learning shows more risky moves (taking turns really close to walls) while in Sarsa we see a smoother general behaviour. The major difference between Sarsa and Q-Learning, is that the maximum reward for the next state is not necessarily used for updating the Q-values (learning table). Instead, a new action, and therefore reward, is selected using the same policy that determined the original action \cite{sarsavsqlearn2}. This is how Sarsa is able to take into account the control policy of the agent during learning. It means that information needs to be stored longer before the action values can be updated, but also means that our robot is going to take risky actions much frequently  \cite{sarsavsqlearn}. This smoother behaviour where forward actions are being exploited in straight tracks leads to higher maximum cumulated rewards. We get values near 3500 in Sarsa while just get cumulated rewards around 2500 in Q-Learning. Running Sarsa for more episodes will cause to get higher average rewards.\\
\newline
The table below provides a numerical comparison of Q-Learning and Sarsa representing the average reward value over 200 consecutive episodes. From the data, one can tell that learning occurs much faster using the Q-Learning technique:

\begin{table}[h!]
\begin{center}
    \label{table:benchmark}
    \begin{tabular}{ l | c | c | }
    \textbf{Episode interval} & \textbf{Q-Learning}  & \textbf{Sarsa}\\ \hline
    0-200 & -114 & -124\\ \hline
    200-400 & -79 & -98\\ \hline
    400-600 & 72 & -75\\ \hline
    600-800 & 212 & -43\\ \hline
    800-1000 & 239 & -43\\ \hline
    1000-1200 & 282 & -6\\ \hline
    1200-1400 & 243 & 55\\ \hline
    1400-1600 & 439 & 65\\ \hline
    1600-1800 & 676 & 104\\ \hline
    1800-2000 & 503 & 127\\ \hline
    2000-2200 & 510 & 361\\ \hline
    2200-2400 & 345 & 164\\ \hline
    2400-2600 & 776 & 698\\ \hline
    2600-2800 & 805 & 550\\ \hline
    2800-3000 & 685 & 240\\
    \end{tabular}
    \caption{Average reward value over a 200 episode interval in 3000 episode long tests using Q-Learning and Sarsa.}
    \label{table:benchmark}
\end{center}
\end{table}

\underline{Variance}. Although the variance presented in Figures \ref{fig:qlearnplot} \& \ref{fig:sarsaplot} is high\footnote{common thing in this scenario}, the averaged plots and the table above show that out robot has learnt to avoid obstacles using both algorithms.\\
\newline
Iterating 2000 episodes more will not make the variance disappear, as we have almost reached the best behaviour possible in this highly discretized environment. Laser values are discretized so that the learning does not take too long. We take only 5 readings with integer values, which are taken uniformly from the 270º lasers horizontal field of view. To sum up, using a simple reinforcement learning technique and just a LIDAR as an input, we get quite decent results.\\
\newline

\newpage

\section{Future directions}
The presented toolkit could be further improved in the following directions:

\begin{itemize}[leftmargin=.2in]
\item Support more autopilot solutions besides APM such as PX4 or Paparazzi.
\item Speed up simulation for robots using autopilots. Currently, due to limitations of the existing implementation, the simulation is set to normal(real) speed.
\item Pull apart environments and agents. Testing different robots in different environments (not only the ones built specifically for them) would make the toolkit more versatile.
\item Provide additional tools for comparing algorithms.
\item Recommendations and results in \emph{mental reharsal} using the presented toolkit.
\end{itemize}





\begin{acknowledgments}

This research has been funded by Erle Robotics.




\end{acknowledgments}








\begingroup
\setlength{\emergencystretch}{10em}
\printbibliography
\endgroup

\end{article}

\end{document}